\documentclass{SCIS2018}

\begin{document}
	
	\ArticleType{RESEARCH PAPER}
	\Year{2018}
	\Month{January}
	\Vol{61}
	\No{1}
	\DOI{}
	\ArtNo{}
	\ReceiveDate{}
	\ReviseDate{}
	\AcceptDate{}
	\OnlineDate{}
	
	\title{Locality preserving projection on SPD
		matrix Lie group: algorithm and analysis}{Locality Preserving Projection on SPD Matrix Lie Group}
	
	\author[1,2]{Yangyang LI}{{liyangyang12@mails.ucas.ac.cn}}
	\author[1]{Ruqian LU}{}
	
	\AuthorMark{Yangyang LI}
	
	\AuthorCitation{Yangyang LI, Ruqian LU}
	
	
	\address[1]{Academy of Mathematics and Systems Science Key Lab of MADIS\\
		Chinese Academy of Sciences, Beijing {\rm 100190}, China}
	\address[2]{University of Chinese Academy of Sciences, Beijing {\rm 100049}, China}
	
	\abstract{Symmetric positive definite (SPD) matrices used as feature descriptors in image recognition are usually high dimensional. Traditional manifold learning is only applicable for reducing the dimension of high-dimensional vector-form data. 
	For high-dimensional SPD matrices, directly using manifold learning algorithms to reduce the dimension of matrix-form data is impossible. The SPD matrix must first be transformed into a long vector, and then the dimension of this vector must be reduced. However, this approach breaks the spatial structure of the SPD matrix space. To overcome this limitation, we propose a new dimension reduction algorithm on SPD matrix space to transform high-dimensional SPD matrices into low-dimensional SPD matrices. Our work is based on the fact that the set of all SPD matrices with the same size has a Lie group structure, and we aim to transform the manifold learning to the SPD matrix Lie group. We use the basic idea of the manifold learning algorithm called locality preserving projection (LPP) to construct the corresponding Laplacian matrix on the SPD matrix Lie group. Thus, we call our approach Lie-LPP to emphasize its Lie group character. We present a detailed algorithm analysis and show through experiments that Lie-LPP achieves effective results on human action recognition and human face recognition.}
	
	
	\maketitle

	\section{Introduction}
	Image recognition, including dynamic action recognition and static image recognition, is a popular research subject in the field of machine vision and pattern recognition \cite{1} \cite{2} \cite{3} \cite{13}. This technique has a wide range of applications in many fields, such as intelligent video retrieval and perceived interaction. One key step of image recognition is to construct a high-quality image feature descriptor, which determines the accuracy rate of recognition. The original image feature descriptor is a pixel matrix, which is usually transformed into a high-dimensional row feature vector because image recognition on the pixel matrix space is difficult. Given the high dimension of the feature vector, manifold learning algorithms are applied to implement image recognition \cite{1} \cite{9} \cite{14} \cite{15}. However, the vector-form feature descriptor breaks the geometric structure of the pixel matrix space and is highly sensitive to various factors such as illumination intensity, background, and object location. To avoid these disadvantages, tensor space dimensionality reduction based on locality preserving projection (LPP) \cite{5} was proposed in \cite{6}; this approach is linear and deals with the image pixel matrix directly. F. Porikli et al. \cite{11} presented a new feature descriptor by computing a feature covariance matrix within any size region in an image, which preserves the local geometric structure of the pixel matrix (see Appendix A in the supplementary file). Covariance matrix, which is also called symmetric positive definite (SPD) matrix descriptor, has mainly been used in static image recognition \cite{10} \cite{11} \cite{13} \cite{28}. For human action recognition, A. Sanin et al. \cite{4} proposed a new method based on spatial-temporal covariance descriptors. H. Tabia et al.\cite{12} applied SPD matrices as descriptors of 3D skeleton location by constructing covariance matrices on 3D joint locations.
	
	A set of SPD matrices with the same size forms a Riemannian manifold. This SPD Riemannian manifold has a group structure that forms an SPD matrix Lie group $S_+^D$. The group operation in \cite{7} is shown as
	$$
	S_1,S_2\in S_+^D, S_1\odot S_2 \doteq exp\left(log\left(S_1\right)+log\left(S_2\right)\right).
	$$
	We proposed that the dimensions of SPD matrices calculated from image regions are especially high in general. As a result of the matrix-form data of the matrix Lie group, directly reducing the dimension of high-dimensional matrix Lie group is difficult without breaking the structure of the matrix form.
	
	Harandi et al. \cite{17} proposed to learn a kernel function to first map the SPD matrices into a higher-dimensional Euclidean space and then use LPP \cite{5} to reduce their dimension. However, this method would distort the geometric and algebraic structure of the SPD manifold and would lose a considerable amount of important structure information. To overcome this limitation, Harandi et al. \cite{18} suggested mapping the original SPD manifold to a Grassmann manifold and then solving the dimension reduction problem on the latter. However, this method has a high time cost. A similar study was conducted by Huang et al. \cite{7}, who proposed transforming SPD matrices to their corresponding tangent space and learning a linear dimensionality reduction map on that tangent space. However, this algorithm needs several parameters, which are sensitive factors that influence the algorithm. Overall, all three methods \cite{7} \cite{17} \cite{18} require a linear space and rely on nonlinear dimensionality reduction mappings.
	
	According to the definition of covariance matrix in \cite{8} \cite{11}, each covariance matrix can be represented by the product of a set of feature vectors. Covariance matrix summarizes the linear bivariate relationships among a set of variables. Thus, we can solve the dimensionality reduction problem directly on the SPD matrix Lie group. We extend the idea of LPP \cite{5} to the dimensionality reduction learning on the SPD matrix Lie group in a new approach called Lie-LPP.\\
	\\
	The main contributions of our work can be summarized as follows:
	\begin{itemize}
		\item  The LPP algorithm in \cite{5} is extended to Lie-LPP and applied to the SPD matrix Lie group. A bilinear dimensionality reduction mapping on the SPD matrix Lie group is obtained, which preserves the intrinsic geometric and algebraic structure of the SPD matrix Lie group.
		\item To overcome the limitation of other methods regarding dimensionality reduction of the SPD matrix Lie group, our method solves the dimensionality reduction problem of the SPD matrix Lie group directly without mapping to other spaces and without needing numerous sensitive parameters, thereby resulting in a simple and straightforward approach.
		\item The graph Laplacian matrix is constructed on the SPD matrix Lie group to reflect the intrinsic geometric and algebraic structure of the original SPD matrix Lie group.
		\item A detailed algorithm analysis in theory is given to analyze the difference and relationship between Lie-LPP and LPP. The main conclusions are shown in Theorems 4.1 and 4.2 and in Proposition 4.1.
	\end{itemize}
	
	This paper is a further extension of the algorithm and analysis in our previous paper \cite{22}. In \cite{22}, we simply presented a simple description of our algorithm. In this paper, we present the detailed algorithm analysis in theory, as well as full experiments on human action and face databases. 
	
	\section{Background}
	
	\subsection{Covariance Matrix}
	
	Suppose a set of $n$ feature vectors from an image region are expressed as the following matrix $F$, where the dimensionality of each feature vector is $d$:
	$$F=\left(f_1,f_2,\dots,f_n\right),$$ 
	$f_i=\left(f_{i_1},f_{i_2},\dots,f_{i_d}\right)$
	is the $i^{th}$ feature vector. The corresponding covariance matrix $C$ with respect to these feature vectors is defined as \cite{8}
	\begin{equation}
	C=\frac{1}{n} FF^T=\frac{1}{n}\sum_{i=1}^{n}f_i f_i^T.
	\end{equation}
	In this definition, we assume the expectation of feature vectors is zero and each term $f_if_i^T$ in the summation is the outer product of the feature vector $f_i$. Obviously, covariance matrix is a semi-positive definite matrix. In this paper, we consider only the positive covariance matrix called the SPD matrix. When the feature vectors are adjacent, the corresponding covariance matrices are also adjacent. Thus, the covariance matrices preserve the local geometric structure of the corresponding feature vectors. The detailed proof is shown in Appendix A.
	
	Covariance matrices (SPD matrices) have several advantages as the feature descriptors of images. First, covariance matrices can fuse all the features of images. Second, they provide a way of filtering noisy information such as illumination and the location of the object in the image. In addition, the size of the covariance matrix is dependent on the dimensionality of feature vectors other than the size of the image region. Thus, we can construct the covariance matrices with the same size from different regions.
	
	\subsection{Geometric structure of SPD Matrix Lie group}
	In machine learning, we usually have to learn an effective metric for comparing data points. In particular, in the image recognition step, a metric is required to measure the distance between two different image feature descriptors. In this paper, we use the SPD matrices as feature descriptors of images. Thus, the corresponding Riemannian metric needs to be constructed on the SPD matrix Lie group to compute the intrinsic geodesic distances between any two SPD matrices. 
	
	The SPD matrix Lie group that we consider in this paper is represented by $\mathcal{S}_+^D$, where every point $S_1 \in \mathcal{S}_+^D$ is a $D \times D$ size matrix. The tangent space of $\mathcal{S}_+^D$ at the identity is \textsl{Sym(D)}, \textsl{a bilinear space of symmetric matrices}. The learned lower-dimensional SPD matrix Lie group is represented by $\mathcal{S}_+^d$. 
	The family of all scalar products on all tangent spaces of SPD matrix Lie group is known as the Riemannian metric. The geodesic distance $d_G\left(S_1,S_2\right)$ between any two points $S_1,S_2 \in \mathcal{S}_+^D$ can be computed under this Riemannian metric. In this paper, we choose the Log-Euclidean metric (LEM) from \cite{21} as the Riemannian metric of SPD matrix Lie group. The detailed definition of LEM is shown in Definition 2.1.
	
	\textbf{Definition 2.1. (Log-Euclidean Metric)} \cite{21} \textsl{The Riemannian metric at a point $S_1\in \mathcal{S}_+^D$ is a scalar product defined in the tangent space $T_{S_1}\left(\mathcal{S}_+^D\right)$:}
	
	\begin{equation}
	\langle T_1,T_2\rangle = \langle D log. T_1, D log. T_2\rangle,
	\end{equation}
	\textsl{where $T_1,T_2 \in T_{S_1}\left(\mathcal{S}_+^D\right)$, $Dlog.T_1$ indicates the directional derivative of the matrix logarithm along $T_1$ \cite{7}.}
	
	LEM is a bi-invariant metric defined on the SPD matrix Lie group. The corresponding theoretical conclusion is shown in \cite{21}.
	
	\textbf{Definition 2.2. (Bi-invariant Metric)} \cite{21} \textsl{Any bi-invariant metric on the Lie group of SPD matrices is also called a LEM because it corresponds to a Euclidean metric (EM) in the logarithm domain.}\\
	The logarithm domain is the tangent space of the SPD matrix Lie group.
	
	\textbf{Corollary 2.3. (Flat Riemannian Manifold)} \cite{21} \textsl{Endowed with a bi-invariant metric, the space of SPD matrices is a flat Riemannian space; its sectional curvature is null everywhere.}
	
	Thus, under LEM, $\mathcal{S}_+^D$ is a flat manifold and locally isometric to the tangent space \textsl{Sym(D)}. In a local neighborhood, the mapping between the SPD matrix Lie group and the corresponding tangent space is represented by the exponential map, and the inverse map is the logarithm shown in Eqs. 3 and 4.
	\begin{align}
	& exp_{S_1}\left(T_1\right)=exp\left(log\left(S_1\right)+D_{S_1}log\cdot T_1\right),\\&
	log_{S_1}\left(S_2\right) = D_{log\left(S_1\right)}exp\cdot\left(log\left(S_2\right)-log\left(S_1\right)\right),
	\end{align}
	where the exponential map is defined at point $S_1 \in \mathcal{S}_+^D$, $T_1 \in T_{S_1}\mathcal{S}_+^D$ a tangent vector. $exp_{S_1}\left(T_1\right)$ is the exponential representation of $T_1$ on $\mathcal{S}_+^D$, and the corresponding logarithmic representation of $S_2$ at $T_{S_1}\mathcal{S}_+^D$ is $log_{S_1}\left(S_2\right)$.
	
	The geodesic distance between $S_1, S_2$ under LEM is defined as follows \cite{7}:
	\begin{equation}
	d_G\left(S_1,S_2\right) = \langle log_{S_1} \left(S_2\right), log_{S_1} \left(S_2\right) \rangle
	= \parallel log\left(S_1\right)- log\left(S_2\right) \parallel_F^2,
	\end{equation}
	where $\|\cdot\|_F$ represents the Frobenius norm of the matrix and $log_{S_1}\left(S_2\right)$ is the logarithm of $S_2$ at $S_1$.
	
	Under LEM, the SPD Matrix Lie group is a complete manifold. Thus, every two points on the SPD matrix Lie group are linked by the shortest geodesic line.
	
	\section{Algorithm}
	In this section, we first analyze the LPP algorithm. Then, we give the construction process of graph Laplacian matrix on the SPD matrix Lie group. Finally, we describe the process of our proposed dimensionality reduction algorithm Lie-LPP. 
	
	\subsection{Locally Preserving Projection (LPP)}
	
	LPP aims to learn a \textsl{linear} dimensionality reduction map to reduce the dimension of high-dimensional vector-form data points, which can be seen as a linear approximation of LEP \cite{16}. This linear dimensionality reduction map optimally preserves the local neighborhood geometric structure of a dataset by building a graph Laplacian matrix on the dataset. The graph Laplacian matrix is a discrete approximation of a differential Laplace operator that arises from the manifold. Let $X=[x_1,x_2,\cdots,x_N] \subset \mathbb{R}^D$ is the input data set distributed on a $d$-dimensional manifold $\mathcal{M}$, which is embedded into $\mathbb{R}^D$. $a$ is the linear dimensionality reduction map. The learned lower-dimensional data set is $Y=a^TX=[y_1,y_2,\cdots,y_N] \subset \mathbb{R}^d$.\\
	\\
	The algorithm of LPP is stated as follows:
	\begin{itemize}
		\item \textbf{Constructing the adjacency graph}: Denote $G$ a graph with $N$ nodes. If $x_i$ and $x_j$ are "close", then a connection exists between nodes $i$ and $j$. The "closenes" between two nodes is measured by the $K$-nearest neighbor method. 
		\item \textbf{Choosing the weights}: Here, the authors denote matrix $W$ as the corresponding weight matrix, where $W$ is a sparse symmetric matrix with size $N \times N$. The element $W_{ij}$ is defined as follows: \\
		$W_{ij}=e^{-\frac{\|x_i-x_j\|^2}{t}}$, if nodes $i$ and $j$ are connected,\\
		$W_{ij}=0$, if nodes $i$ and $j$ are not connected.
		\item \textbf{Eigenmaps}: The following generalized eigenvector problem is solved to obtain the corresponding dimension reduction map $a$:
		\begin{equation}
		XLX^Ta=\lambda XDX^Ta,
		\end{equation}
		where $D$ is a diagonal matrix, $D_{ii}=\sum_{j} W_{ji}$. $L=D-W$ is the graph Laplacian matrix. 
	\end{itemize}
	
	Thus, the critical step of LPP is to construct the graph Laplacian matrix $L$ on data points. The global lower dimensional representations are learned by solving the corresponding generalized eigenfunction.
	
	\subsection{Laplace operator on SPD matrix Lie group}
	The Laplace operator is a significant operator defined on Riemannian manifold \cite{27}. It measures the intrinsic structure of manifold such as the curvature of manifold, the similarities among different points on Riemannian manifold. The Laplacian matrix on a graph is a discrete analog of the Laplace operator that we are familiar with in functional analysis \cite{26}. The critical step of the LPP algorithm is to construct a graph Laplacian matrix to represent the intrinsic local geometric structure of data points, which is applied only on vector-form data points. For Lie-LPP, we aim to uncover the intrinsic structure of SPD matrices, which are essentially different from vector-form data points in a spatial structure. The Laplace operator is defined based on the Riemannian metric. For vector-form data points, the Laplacian matrix is constructed based on the EM in each local patch. For SPD matrices, we need to use LEM \cite{29} \cite{21} to construct the corresponding Laplacian matrix. 
	
	Several difficulties are encountered for the learning of the Laplacian matrix on SPD matrices. One critical difficulty is the discrete representations of the first- and second-order derivatives on SPD matrix space. Another difficulty is the need to find the representation of the Laplace operator on the SPD matrix Lie group, which is different from the structure on vector space. To construct a Laplacian matrix on SPD matrices, we need to solve these difficulties during the learning process. In addition, the core of Lie-LPP is to construct an accurate Laplacian matrix on the SPD matrix Lie group before dimensionality reduction. The construction process of a Laplacian matrix on SPD matrices and the descriptions for solving these difficulties are given in detail in a separate section. 
	
	To better understand the construction of the Laplacian matrix, we first present an intuitive example of a set of one dimensional nodes. Consider a graph $G$ with $n$ nodes. Every node $i$ is adjacent to two nodes $i-1$ and $i+1$. If we assign value $v_i$ to node $i$, then the Laplacian is represented as $\left(v_{i+1}-v_i\right)-\left(v_i-v_{i-1}\right)$. Thus, $v_{i+1}-v_i$ is the discrete analog for a first-order derivative defined over the real number line.  $-2\left(v_i-\frac{1}{2}v_{i-1}-\frac{1}{2}v_{i+1}\right)$ is the discrete approximation of the second-order derivative.
	For higher dimensions the 'normalized' graph Laplacian $\Delta_{nm}$ about function $f$ is defined as:
	\begin{equation}
	\left(\Delta_{nm}f\right)\left(i\right)=f\left(i\right)-\frac{1}{deg\left(i\right)}\sum_{j=1}^{n}W_{ij}f\left(j\right),
	\end{equation} 
	where the degree function $deg$ is defined as $deg\left(i\right)=\sum_{j=1}^{n}W_{ji}$. $W_{ji}$ is the heat kernel weight defined in the same way as in the second algorithmic procedure of LPP \cite{5}.
	
	The abovementioned Laplacian matrix is defined on vector-form data points only. In the following, we construct the 'normalized' graph Laplacian matrix on the SPD matrix Lie group endowed with LEM. Suppose a parameterized SPD matrix Lie group $\Sigma$ defined as $\Sigma: \mathbb{R}^d \rightarrow \mathcal{S}_+^D$, we call the vector $\overrightarrow{\Sigma\left(x\right)\Sigma\left(x+u\right)}$ as the standard first-order derivative on $\mathcal{S}_+^D$ \cite{20}:
	\begin{equation}
	\begin{split}
	\overrightarrow{\Sigma\left(x\right)\Sigma\left(x+u\right)}=\Sigma\left(x\right)^\frac{1}{2}\left(log\Sigma\left(x+u\right)-log\Sigma\left(x\right)\right)\Sigma \left(x\right)^\frac{1}{2}.
	\end{split}
	\end{equation}  
	The Laplace-Beltrami operator $\Delta$ about function $\Sigma$ is defined as:
	\begin{equation}
	\begin{split}
	&\Delta \Sigma = \sum_{i=1}^{d}\Delta_i \Sigma,\\
	& \Delta_i \Sigma=\partial_i^2\Sigma-2\left(\partial_i\Sigma\right)\Sigma^{\left(-1\right)} \left(\partial_i\Sigma\right).
	\end{split}
	\end{equation}
	
	To approximate the graph Laplacian matrix on SPD matrices, we first need to approximate the first- and second-order derivatives on SPD matrix Lie group. Under the approximation of the second-order derivative, the corresponding approximation of the Laplace-Beltrami operator on $\mathcal{S}_+^D$  \cite{20} is:
	\begin{equation}
	\Delta_u \Sigma=\partial_u^2\Sigma-2\left(\partial_u \Sigma\right)\Sigma^{\left(-1\right)}\left(\partial_u\Sigma\right)
	=\overrightarrow{\Sigma\left(x\right)\Sigma \left(x+u\right)}+\overrightarrow{\Sigma\left(x\right)\Sigma\left(x-u\right)}+O\left(\|u\|^4\right).
	\end{equation}
	To compute the complete Laplacian matrix of the SPD matrix Lie group of Eq. 9, we only have to compute the Laplace operator along $d$ orthonormal directions.
	
	For a discrete dataset, suppose $\Sigma_1, \Sigma_2,\cdots,\Sigma_N $ are a set of SPD matrices generated from $\Sigma$. The ?normal' graph Laplacian matrix on this data set is
	\begin{equation}
	\left(\Delta_{nm}\Sigma_i\right)=\Sigma_i^\frac{1}{2}\left(log\left(\Sigma_i\right)-\sum_{j=1}^{n}\widetilde{W}_{ij} log\left(\Sigma_j\right)\right)\Sigma_i^\frac{1}{2},
	\end{equation}
	where $\widetilde{W}_{ij}=e^{-\frac{\|log\left(\Sigma_i\right) -log\left(\Sigma_j\right)\|_F^2}{t}}$, if $i$ and $j$ are connected, else $\widetilde{W}_{ij}=0$. The corresponding graph Laplacian matrix on a set of SPD matrices is $\widetilde{L}=\widetilde{D}-\widetilde{W}$, where $\widetilde{W}$ is a symmetric matrix, $\widetilde{W}_{ij}$ defined as above and $\widetilde{D}$ is a diagonal matrix, $\widetilde{D}_{ii}=\sum_{j}\widetilde{W}_{ji}$. $\widetilde{L}$ is a discrete representation of the Laplace operator on the SPD matrix Lie group.
	
	\subsection{Lie-LPP Algorithm}
	
	On the basis of the definition of graph Laplacian matrix on $\mathcal{S}_+^D$ and the algorithmic procedures of LPP, we show our Lie-LPP algorithm as follows:
	\subsubsection{Lie-LPP Algorithm}
	For the SPD matrix Lie group $\mathcal{S}_+^D$, the SPD matrix logarithms in the tangent space are also symmetric matrices. The bilinear mapping between tangent spaces is defined as follows:
	\begin{equation}
	f\left(log \left(S_1\right) \right)= A^T log \left(S_1\right) A.
	\end{equation}
	The corresponding mapping between SPD matrix Lie groups is
	\begin{equation}
	g\left(S_1\right) = exp \circ f\left(log\left(S_1\right)\right)=exp\left(A^T log\left(S_1\right) A\right),
	\end{equation}
	where $S_1 \in \mathcal{S}_+^D$, $A$ is a linear map matrix, $f$ is the corresponding map defined on Lie algebras, and $g$ is the derived map defined on SPD matrix Lie groups.
	
	$g \left(S_1\right)$ is still an SPD matrix; this idea is easily proven. In this paper, we attempt to learn a transformation matrix $A$ where $A\in \mathcal{R}^{D\times d}$ is a full column rank matrix. $D \times D$ is the size of $\mathcal{S}_+^D$, and $d \times d$ is the mapped size of $\mathcal{S}_+^d$, $D>>d$. The linear map matrix $A$ is proven to preserve the algebraic structure of $\mathcal{S}_+^D$. To obtain a discriminative SPD matrix Lie group $\mathcal{S}_+^d$, $A$ should also inherit and preserve the geometric structure of $\mathcal{S}_+^D$. According to the idea of LPP, the key step of the Lie-LPP algorithm is to construct the Laplacian matrix $L$ on $\mathcal{S}_+^D$, which reflects the local geometrical structure of $\mathcal{S}_+^D$. Under LEM, $\mathcal{S}_+^D$ is locally isometric to the tangent space of $\mathcal{S}_+^D$. Thus, the geodesic distance between two SPD matrices is equal to the Euclidean distance between the corresponding points on tangent space.
	
	Suppose the input data points are $S_1,S_2,\ldots, S_N \in \mathcal{S}_+^D$, and the output sample points are $Y_1,Y_2, \dots, Y_N \in \mathcal{S}_+^d$, where $N$ is the number of sampled points.
	\\
	\\
	\textbf{The algorithm steps of Lie-LPP are as follows:}
	\begin{itemize}
		\item The first step is to divide the input SPD matrices into a set of local patches. We use the $K$-nearest method to find the $K$-nearest neighborhoods $U_i$ of every point $S_i$, where the distance metric between two points is defined as their geodesic distance on the SPD matrix Lie group $\mathcal{S}_+^D$ using Eq. 5.
		\item The second step is to construct a weight matrix $W$ on each local patch $U_i$ to represent the local intrinsic geometric structure of $U_i$.\\
		
		\quad \quad $\widetilde{W}_{ij}=e^{-\frac{\parallel log\left(S_i\right)-log\left(S_j\right) \parallel _F^2}{t}}$, if $S_j \in U_i$, 
		
		\quad \quad $\widetilde{W}_{ij}=0$, else if $S_j \notin U_i$.\\
		
		The definition of weight value $\widetilde{W}_{ij}$ is based on the construction of a Laplacian matrix on input SPD matrices from Eq. 11.
		\item The third step is to compute the eigenvalues and the corresponding eigenvectors for the generalized eigenfunction problem
		\begin{equation}
		S^T\widetilde{L}SA=\lambda S^T\widetilde{D}SA,
		\end{equation}
		where $S=[log\left(S_1\right),log\left(S_2\right),\cdots,log\left(S_N\right)]^T$ is a partitioned matrix.
	\end{itemize}
	
	
	\subsubsection{Optimal Embedding}
	The optimal dimension reduction map $A$ is obtained by minimizing the following energy function: 
	\begin{equation}
	\frac{1}{2}\sum_{i,j} d_G \left(Y_i, Y_j\right) \widetilde{W}_{ij},
	\end{equation}
	where $d_G \left(Y_i, Y_j\right)$ is the geodesic distance between $Y_i$ and $Y_j$, while $\widetilde{W}_{ij}$ is the corresponding weight.
	
	According to the definition of geodesic distance and LEM on the SPD matrix Lie group in Eq. 5, the energy function Eq. 15 can be transformed into the following equation: 
	\begin{equation}
	\frac{1}{2} \sum_{i,j} \parallel log Y_i - log Y_j \parallel _F^2 \widetilde{W}_{ij},
	\end{equation}
	According to Eq. 12, the optimization function Eq. 16 is represented as follows:
	\begin{equation}
	\begin{split}
	& \frac{1}{2} \sum_{i,j} \parallel logY_i-log Y_j \parallel_F^2 \widetilde{W}_{ij} \\
	& = \frac{1}{2} \sum_{i,j}\parallel A^TlogS_iA-A^TlogS_jA\parallel_F^2 \widetilde{W}_{ij}\\
	& = tr\left(P^TS^T\left(\widetilde{D}-\widetilde{W}\right)SP\right)\\
	& =tr\left(P^TS^T\widetilde{L}SP\right),
	\end{split}
	\end{equation}
	where $\widetilde{W}$ and $\widetilde{D}$ are two sparse $N\times N$ block matrices, $\widetilde{W}=[\widetilde{W}_{ij} I_{D\times D}]$, $\widetilde{D}=diag\left(\widetilde{D}_{11}I_{D\times D},\cdots, \widetilde{D}_{NN}I_{D\times D}\right)$. $P=A A^T \in R^{D \times D}$ and $\widetilde{L}=\widetilde{D}-\widetilde{W}$ are all semi-SPD matrices. Thus, $P^T S^T \left(\widetilde{D}-\widetilde{W}\right) S P$ is also a semi-SPD matrix, and its eigenvalues are all non-negative. We have $tr\left(P^T S^T \widetilde{L} SP\right) \geq 0$. To compute the minimum value of Eq. 17, we just need to compute the minimum eigenvalues of matrix $P^T S^T \widetilde{L} S P$.\\
	To avoid obtaining a singular solution, we impose a constraint as follows:
	\begin{equation}
	P^T S^T \widetilde{D} S P = I.
	\end{equation}
	Then, the corresponding minimization problem turns to
	\begin{equation}
	\begin{split}
	min\quad & tr(P^T S^T \widetilde{L} S P),\\
	s.t. \quad & P^T S^T \widetilde{D} S P=I.
	\end{split}
	\end{equation}
	We use the Lagrange multiplier method to solve the minimization problem
	\begin{equation}
	\begin{split}
	L\left(A, \lambda \right) &= tr \left(P^T S^T \widetilde{L} S P\right)- \lambda \left(tr \left(P^T S^T \widetilde{D} S P-I\right)\right)\\
	& = tr \left(AA^T\widetilde{L}SAA^T-\lambda\left(AA^TS^T\widetilde{D}SAA^T-I\right)\right).
	\end{split}
	\end{equation}
	As $L$'s derivative to $A$, we obtain the following:
	\begin{equation}
	\frac{\partial L\left(A,\lambda\right)}{\partial A} = 4tr\left(A^T S^T \widetilde{L} S A A^T - \lambda A^T S^T \widetilde{D} S A A^T\right),
	\end{equation}
	where $P=A A^T$ and $S^T \widetilde{L} S$ are both semi-SPD matrices,
	$tr\left(S^T \widetilde{L} S A A^T\right)= tr\left(A A^T S^T \widetilde{L} S\right)$. Eq. 21 can be derived as 
	\begin{equation}
	\frac{\partial L\left(A,\lambda\right)}{A}=4 tr\left(A^TAA^TS^T\widetilde{L}S-\lambda A^TAA^TS^T\widetilde{D}S\right)
	= 4 tr\left(\left(S^T\widetilde{L}SA-\lambda S^T\widetilde{D}SA\right)A^TA\right).
	\end{equation}
	$A$ is a $D \times d$ full rank matrix; thus, $A^TA$ is a $d \times d$ SPD matrix. To obtain $\frac{\partial L\left(A, \lambda\right)}{\partial A}=0$, we need to minimize the following generalized eigenfunction problem:
	\begin{equation}
	S^T\widetilde{L}SA=\lambda S^T\widetilde{D}SA.
	\end{equation}
	
	We obtain the bottom smallest $d$ eigenvalues $0\leq \lambda_1 \leq \lambda_2 \leq \ldots \leq \lambda_d$, and the corresponding eigenvectors $a_1,a_2,\ldots, a_d\in \mathbb{R}^{D \times 1}$. $A=\left[a_1,a_2,\ldots,a_d\right] \in \mathbb{R}^{D \times d}$ is the learned linear dimensionality reduction map matrix. The corresponding dimension reduction map $g$ between $\mathcal{S}_+^D$ and $\mathcal{S}_+^d$ is
	\begin{equation}
	Y_i=g\left(S_i\right)=exp\left(A^Tlog\left(S_i\right)A\right).
	\end{equation}
	
	The lower-dimensional SPD matrix Lie group $\mathcal{S}_+^d$ preserves the local geometric and algebraic structure of $\mathcal{S}_+^D$, which is maintained by the Laplacian matrix $L$ on $\mathcal{S}_+^D$. Good similarity between two points corresponds to a large weight between them. In addition, through the construction of a graph Laplacian matrix on the SPD matrices, the reduction map is learned based on global data points. The Laplacian matrix can be viewed as an alignment matrix that aligns a set of local patch structures to obtain global lower-dimensional representations by solving a generalized eigenfunction. Unlike other methods \cite{7} \cite{17} \cite{18}, our method uncovers the intrinsic structure of SPD matrices by constructing this discrete Laplacian matrix without the help of other spaces.
	
	\section{Algorithm Analysis}
	
	In this section, we mainly analyze the relationships between the proposed Lie-LPP and LPP \cite{5} in theory. In the first subsection, we present dimension reduction error comparisons between these two algorithms. In the second subsection, we analyze the similarity relation between them.
	
	\subsection{Comparison with LPP} 
	We analyze the reconstruction errors during dimension reduction of Lie-LPP and LPP from two aspects. First, we analyze the local weight matrix construction. Second, we analyze the global alignment matrix and the null space of the alignment matrix. The dimension reduction losses of the two algorithms are determined by the corresponding graph Laplacian matrices defined on data points. In this section, we mainly compare Lie-LPP and LPP in theory to analyze the improvements of our algorithm.
	
	First, we show the relationship between two graph Laplacian matrices, which are defined on vector-form data points and SPD matrix-form data points.
	
	The local weight matrix $W$ of LPP is defined in the second step of the LPP algorithm:
	\begin{equation*}
	W_{ij}=e^{-\frac{\|x_i-x_j\|^2}{t}}, 
	\end{equation*}
	where $x_i$ and $x_j$ are in the same neighborhood.
	
	The local weight matrix $\widetilde{W}$ of Lie-LPP is defined in the second step of the Lie-LPP algorithm
	\begin{equation*}
	\widetilde{W}_{ij}=e^{-\frac{\parallel log\left(S_i\right)-log\left(S_j\right) \parallel _F^2}{t}},
	\end{equation*}
	where $S_i$ and $S_j$ are also in the same neighborhood.
	
	The distance $\|x_i-x_j\|$ between $x_i$ and $x_j$ is computed based on EM. However, the distance $\parallel log\left(S_i\right)-log\left(S_j\right) \parallel_F$ between $S_i$ and $S_j$ is computed based on LEM. EM is not the real Riemannian metric of the embedded manifold $\mathcal{M}$ defined in the LPP algorithm. We mentioned in Subsection 2.2 that LEM is the intrinsic Riemannian metric defined on the SPD matrix Lie group. Thus, the distance $\|x_i-x_j\|$ under EM is not the intrinsic geodesic distance $dist_\mathcal{M}\left(x_i,x_j\right)$ on $\mathcal{M}$ and obviously $dist_\mathcal{M}\left(x_i,x_j\right) \geq \|x_i-x_j\|$. Under LEM, the intrinsic geometric structure of SPD matrix Lie group can be determined, and $\parallel log\left(S_i\right)-log\left(S_j\right) \parallel_F$ is the real geodesic distance between $S_i$ and $S_j$. To our knowledge, $x_i$ and $S_i$ are two different feature descriptors of the same image in computer vision. In the Appendix, we prove that the SPD matrix descriptors preserve the geometric structures of vector-form feature descriptors. Thus, we have 
	$$\parallel log\left(S_i\right)-log\left(S_j\right) \parallel_F^2 \geq \|x_i-x_j\|^2.$$ Under this analysis, we have
	\begin{equation}
	W_{ij} \geq \widetilde{W}_{ij}.
	\end{equation}
	
	The corresponding graph Laplacian matrices defined on $\{x_i\}$ and $\{S_i\}, i=1,2,\cdots,N$ are represented as $L$ and $\widetilde{L}$, respectively. On the basis of the above analysis, we present our first comparison conclusion in Theorem 4.1.\\
	\\
	\textbf{Theorem 4.1.} If the datasets $\{x_i\}$ and $\{S_i\}, \left(i=1,2,\cdots,N\right)$ are two different feature descriptors of the same $N$ images, then we have $L \succeq \widetilde{L}$, that is,
	\begin{equation*}
	\lambda_i\left(L\right) \geq \lambda_i \left(\widetilde{L}\right),
	\end{equation*}
	for all $i=1,2,\cdots,N$.\\
	\textbf{Proof:} According to the definition of weight matrices $\widetilde{W}$ and $W$, we obtain Eq. 25. Then $D-\widetilde{D}$ is a positive diagonal matrix. We know that $\widetilde{L}=\widetilde{D}-\widetilde{W}$ and $L=D-W$; thus, $L-\widetilde{L} = D-\widetilde{D}-\left(W-\widetilde{W}\right)$ is also a Laplacian matrix. $L-\widetilde{L}$ is a semi-positive definite symmetric matrix. We have
	\begin{equation*}
	\lambda_i(L-\widetilde{L}) \geq 0, ~~ ~~~L \succeq \widetilde{L},
	\end{equation*}
	for all $i=1,2,\cdots,N$.
	
	For two symmetric matrices $L,\widetilde{L}$, if $L-\widetilde{L} \succeq 0$, then we write $L \succeq \widetilde{L}$. If $L \succeq \widetilde{L}$, then 
	\begin{equation*}
	\lambda_i\left(L\right) \geq \lambda_i \left(\widetilde{L}\right),
	\end{equation*} 
	for every $i$.
	
	For a special situation, if the original sub-manifold $\mathcal{M}$ in LPP is highly curved and the Riemannian curvature is not zero everywhere, then the eigenvalues of $L$ are strictly greater than those of $\widetilde{L}$, that is, $\lambda_i\left(L\right) > \lambda_i\left(\widetilde{L}\right)$ for every $i$ under this situation. Therefore, we have proven this theorem.\\
	$\blacksquare$
	
	After analyzing the relationship between $L$ and $\widetilde{L}$ in Theorem 4.1, we need to analyze the dimension reduction errors of Lie-LPP and LPP. In the dimension reduction step, the two algorithms need to minimize the following generalized eigenvalue function:
	\begin{equation*}
	E=\frac{1}{2} \sum_{i,j}\left(y_i-y_j\right)^2 W_{ij} = Y^T L Y,
	\end{equation*}
	where $\{y_1,y_2,\cdots,y_N\}$ are lower-dimensional representations of $\{x_i\}$ or $\{S_i\}$, for $i=1,2,\cdots,N$. 
	
	The dimension reduction errors of Lie-LPP and LPP are measured by the smallest eigenvalues of graph Laplacian matrices. Suppose the dimension reduction error under LPP is represented as $E$ and the error under Lie-LPP is represented as $\widetilde{E}$. On the basis of Theorem 4.1, we present our second comparison conclusion in Theorem 4.2.\\
	\\ 
	\textbf{Theorem 4.2.} The dimension reduction error $\widetilde{E}$ under Lie-LPP is less than the dimension reduction error $E$ under LPP, that is, we have  
	\begin{equation*}
	\|\widetilde{E}\|_F \leq \|E\|_F.
	\end{equation*}
	\textbf{Proof:} According to the algorithm procedures of the Laplacian eigenmap, the dimension reduction errors $E$ of LPP and Lie-LPP are mainly determined by the smallest eigenvalues of the graph Laplacian matrix. Thus, the norm of general reconstruction error $E$ is measured as
	\begin{equation*}
	\|E\|_F = \sum_{i=1}^{d} \lambda_i,
	\end{equation*} 
	where $d$ is the intrinsic dimension of lower-dimensional representations.
	
	From Theorem 4.1, we can deduce that for the same image database, the graph Laplacian matrix $\widetilde{L}$ constructed on the SPD matrices is lower than that on the vector-form descriptor. Thus, we have $\lambda_i\left(L\right) \geq \lambda_i \left(\widetilde{L}\right)$, for all $i=1,2,\cdots,N$. Then, on the basis of the definition of the dimension reduction error norm of Lie-LPP and LPP, we have
	\begin{equation*}
	\|\widetilde{E}\|_F \leq \|E\|_F.
	\end{equation*}
	Under the same special situation in Theorem 4.1, if the embedded manifold $\mathcal{M}$ in LPP is highly curved, then the reconstruction error $E$ is strictly greater than $\widetilde{E}$.
	\begin{equation*}
	\|\widetilde{E}\|_F < \|E\|_F.
	\end{equation*}
	The key reason for this situation is that in LPP, the authors used EM to determine the local intrinsic geometric structure of $\mathcal{M}$, which is not the real local Riemannian metric of $\mathcal{M}$.\\  $\blacksquare$
	
	\subsection{Connection to LPP}
	
	We also present a theoretical analysis of the similarity relation between Lie-LPP and LPP aside from reconstruction error comparisons between these two algorithms. Through the analysis, we can see that under the following special situation, Lie-LPP is equivalent to LPP by defining a new weight matrix. Suppose the vector form descriptor of an object is represented as $x_i$ a row vector. The corresponding SPD matrix descriptor of this object is shown as $S_i=x_i^Tx_i$. SPD matrix Lie group is a flat Riemannian manifold, where the local neighborhood of the Lie group is locally isometric to the corresponding tangent space. Thus, the local tangent space can be approximately represented by the local neighborhood of the SPD matrix Lie group. Under this special situation and theoretical analysis, Lie-LPP can be transformed into LPP with a special weight matrix.
	The theoretical analysis is stated in Proposition 4.1.\\
	\\
	\textbf{Proposition 4.1.} The vector form descriptor of an object be $x_i$ and the corresponding SPD matrix descriptor be $x_i^Tx_i$. Under this special situation, Lie-LPP can be transformed into LPP by defining a new weight matrix.\\
	\\
	\textbf{Proof:} First, we give the following representation of the generalized eigenvalue function of Lie-LPP:
	\begin{equation*}
	\begin{split}
	&S^T \widetilde{L}SA = \lambda S^T \widetilde{D}SA,\\
	&S^T \widetilde{W} SA=\left(1-\lambda\right)S^T\widetilde{D}SA,
	\end{split}
	\end{equation*}
	where $\widetilde{L}=\widetilde{D}-\widetilde{W}$ is the corresponding graph Laplacian matrix defined on a set of SPD matrices $S$ represented as $S^T=[x_1^Tx_1, x_2^Tx_2, \cdots, x_N^Tx_N]$, and $S^T$ is the transpose matrix of $S$.\\
	By rewriting $S^T$ as a matrix representation, we obtain the following representation:
	\begin{equation*}
	S^T=[x_1^T, x_2^T,\cdots, x_N^T]
	\left( \begin{array}{ccc}
	x_1 & \cdots & 0\\
	\vdots & \ddots & \vdots\\
	0 & \cdots & x_N \\
	\end{array}
	\right),
	\end{equation*}\\
	Under this representation, we obtain
	\begin{equation*}
	S^T \widetilde{W} SA=[x_1^T, x_2^T, \cdots, x_N^T]
	\left( \begin{array}{ccc}
	x_1 & \cdots & 0\\
	\vdots & \ddots & \vdots\\
	0 & \cdots & x_N \\
	\end{array}
	\right) \widetilde{W}
	\left( \begin{array}{ccc}
	x_1^T & \cdots & 0\\
	\vdots & \ddots & \vdots\\
	0 & \cdots & x_N^T \\
	\end{array}
	\right) 
	\left( \begin{array}{c}
	x_1 \\
	x_2\\
	\vdots \\
	x_N \\
	\end{array}
	\right)A.
	\end{equation*}
	\\
	Define 
	\begin{equation*}
	W_V=\left( \begin{array}{ccc}
	x_1 & \cdots & 0\\
	\vdots & \ddots & \vdots\\
	0 & \cdots & x_N \\
	\end{array}
	\right) \widetilde{W}
	\left( \begin{array}{ccc}
	x_1^T & \cdots & 0\\
	\vdots & \ddots & \vdots\\
	0 & \cdots & x_N^T \\
	\end{array}
	\right),
	D_V=\left( \begin{array}{ccc}
	x_1 & \cdots & 0\\
	\vdots & \ddots & \vdots\\
	0 & \cdots & x_N \\
	\end{array}
	\right) \widetilde{D}
	\left( \begin{array}{ccc}
	x_1^T & \cdots & 0\\
	\vdots & \ddots & \vdots\\
	0 & \cdots & x_N^T \\
	\end{array}
	\right).
	\end{equation*}\\
	Under this new weight matrix $W_V$, we rewrite $S^T \widetilde{W}SA$ as follows:
	\begin{equation*}
	S^T \widetilde{W}SA=[x_1^T, x_2^T, \cdots, x_N^T] W_V 
	\left( \begin{array}{c}
	x_1 \\
	x_2\\
	\vdots \\
	x_N \\
	\end{array}
	\right)A.
	\end{equation*}
	Suppose  
	$X = \left( \begin{array}{c}
	x_1 \\
	x_2\\
	\vdots \\
	x_N \\
	\end{array}
	\right)
	$, $S^T \widetilde{W} SA= X^T W_V XA$.
	
	Thus, under this new weight matrix $W_V$, Lie-LPP is equivalent to LPP. The above analysis shows that the Lie-LPP algorithm can be transformed into LPP by defining a new weight matrix $W_V$ if the feature descriptors of vector form and SPD matrix form are $\{x_i\}$ and $\{x_i^T x_i\}$, $i=1,2,\cdots,N$, respectively.\\ $\blacksquare$
	
	\section{Experiments}
	In this section, we first report the results we obtained after running Lie-LPP on two human action databases, namely, Motion Capture HDM05 \cite{24} and CMU Motion Graph. We compare our algorithm with traditional manifold learning algorithms through two experiments. In the second part, we test our algorithms on a static face database e.g., extended Yale Face Database B (YFB DB). We compare LEML algorithm \cite{7} and SPD-ML algorithm \cite{18} with our algorithm, after which we present the experimental comparison between Lie-LPP and LPP \cite{5}.
	
	\subsection{Human Action Recognition}
	
	In this subsection, we test Lie-LPP on two human action databases. Each action segment trajectory can be seen as a curve traversing a manifold. Action recognition involves classifying the different action curves. In the recognition step, we use the nearest neighborhood framework to classify human action sequences. After feature extraction, the embedded covariance feature descriptors form an SPD matrix, which belongs to a low-dimensional SPD matrix Lie group. 
	
	\subsubsection{Motion Capture HDM05 Database}
	The HDM05 database \cite{24} contains more than $70$ motion classes in $10$ to $50$ realizations executed by various actors. The actions are performed by five subjects whose names are 'bd','bk', 'dg', 'mm', and 'tr'. For each subject, we choose the following $14$ actions: 'clap above head', 'deposit floor', 'elbow to knee', 'grab high', 'hop on both legs', 'jog', 'kick forward', 'lie down on the floor', 'rotate both arms backward', 'sit down on chair', 'sneak', 'stand up', and 'throw basketball'. We choose three motion fragments per subject per action. Thus, the total number of motion fragments in this experiment is $210$. The dataset provides the $3D$ locations of $31$ joints over time acquired at the speed of $120$ frames per second. In our experiment, we describe each action observed over $T$ frames \cite{3} by its joint covariance descriptor, which is an SPD matrix of size $93 \times 93$. $T$ is a parameter. In this experiment, we choose $T = 20$.
	
	If we directly perform the classification in the original $93\times 93$ dimensional SPD matrix Lie group, the experimental time cost would be as high as up to $21133.9$s. In practice, each joint action is controlled by only a few features, which can be much less than $93$. Therefore, dimension reduction is necessary before recognition. We perform eight groups of experiments. Each experiment is divided into two parts. The first part involves using the use Lie-LPP algorithm to reduce the dimension of SPD matrices, while the second part involves using the leave-one-out method to compute the recognition rate. We measure the similarity between two SPD matrices under two Riemannian metrics, namely, EM and LEM, for comparison. The final comparison results are shown in Table 1. The purpose of this experiment is to analyze the time complexity of Lie-LPP. Thus, we present comparisons of our method under different Riemannian metrics and different reduced dimensions. Under the same reduced dimensions $d\times d$, the accuracy rates under LEM are higher than those under EM overall, but the time cost under LEM is higher than that under EM because of the especially high time cost of the log operation on the SPD matrix. Under the same Riemannian metric, the time cost increases along with the growth of the reduced dimensions. If we do not reduce the dimensionality of SPD matrices before recognition, then the recognition accuracies under both Riemannian metrics are $0.5857$ and $0.6000$, respectively. However, the recognition rates under the low-dimensional SPD matrix Lie group with LEM are $0.7619$, $0.7857$, and $0.7905$ for dimensions reduced $10\times10$, $15\times15$, and $20\times20$, respectively (Table 1); these rates are much higher than they would be without dimension reduction. In conclusion, reducing the dimension of the original SPD matrices is necessary before recognition. Our method reduces the time cost of the experiment and improves the recognition rate. In addition, on the basis of the LEM, the recognition rates are relatively higher. Thus, the intrinsic structure of SPD matrices is more accurately determined based on LEM.  
	
	\begin{table}[tbp]
		\centering  
		\caption{Time comparison, Accuracy rates of HDM05 database under different reduced dimensions and different Riemannian metrics, i.e., EM and LEM.}
		\begin{tabular}{l|c|c|c}  
			\hline
			Methods  &Dimension $d\times d$ &time &Accuracy Rate \\ \hline
			Lie-LPP-EM  &$10 \times 10$ &$16.1961s$ &0.5905\\ 
			Lie-LPP-EM  &$15 \times 15$ &$21.0645s$ &0.6095\\ 
			Lie-LPP-EM & $20 \times 20$ &$30.2843s$ &0.6095 \\ \hline
			Lie-LPP-LEM &$10 \times 10$ &$275.156s$ &0.7619 \\
			Lie-LPP-LEM &$15 \times 15$ &$822.289s$ &0.7857 \\
			Lie-LPP-LEM &$20 \times 20$ &$1003.13s$ &0.7905\\ \hline
			Lie-LPP-EM &$93 \times 93$ &$438.777s$ &0.5857\\
			Lie-LPP-LEM &$93 \times 93$ & $21133.9s$ &0.6000\\ \hline
		\end{tabular}	
	\end{table}  
	
	\subsubsection{CMU Motion Graph Database}
	
	We consider four different action classes, namely, walking, running, jumping, and swing. Each class contains 10 sequences or a total of 40 sequences for our experiment. A total of $31$ joints are marked in the skeletal joint sequences. Only the root joint is represented by a $6D$ vector; the other $30$ joints are represented by $3D$ rotation angle vectors. Each action frame is represented by a $96D$ vector. The SPD matrix, which is a $96 \times 96$ matrix, is constructed by computing the covariance of $T$ frames subwindows, where we choose $T=20$. To guarantee connectivity between subwindows, we take a $\frac{T}{2}$ frames that overlap between adjacent subwindows, as mentioned in \cite{3}. 
	
	In this experiment, we use leave-one-out method to compute the recognition rate. Every time, we choose one sequence of each class as the test set and the remaining sequences as the training set. The comparison results in Table 2 show the recognition accuracies for different action classes, where the recognition accuracy of our method is $0.975$, which is higher than that of the other three methods. Notably, the SPD matrix descriptor under joint locations in our method obtains a surprising result.
	
	\begin{table}[tbp]
		\centering  
		\caption{Classification performance of CMU Motion Graph database together with the comparison results for Lie-LPP and traditional manifold learning algorithms.}
		\begin{tabular}{l|c c}  
			\hline
			Methods   &CMU &Dimension $d$ \\ \hline
			\textbf{Lie-LPP}  &\textbf{0.9750} &$3 \times 3$ \\ 
			LPP \cite{5}  &0.8500 &9 \\ 
			LEP \cite{16} & 0.9000 &9 \\
			PCA \cite{19} &0.2500 &9 \\\hline
		\end{tabular}	
	\end{table}  
	
	\subsection{Human Face Recognition}
	
	In \cite{22}, we tested Lie-LPP on human face databases, namely, Yale Face DB and YTC DB. In this subsection, we further test our algorithm on another static human face database, namely, YFB DB. Only two other algorithms, namely, LEML \cite{7} and SPD-ML \cite{18}, are similar to our method in that they reduce the dimensionality of SPD matrix manifold. In this subsection, we report the comparison results for our method with LEML, SPD-ML, the manifold learning algorithm LPP \cite{5}, and linear dimensionality reduction PCA \cite{19}.
	
	\subsubsection{Extended Yale Face Database B}
	
	YFB DB \cite{25} contains $2414$ single-light-source images of $38$ individuals each seen under approximately $64$ near-frontal images and under different illuminations per individual. For every subject in a particular pose, an image with ambient illumination was captured. The face region in each image is resized to $32 \times 32$. We use the raw intensity feature to construct the corresponding SPD matrix, in which we follow \cite{7} to construct the SPD matrix for each image. For this dataset, because the image size is $32 \times 32$, the size of the corresponding SPD matrix is $1024 \times 1024$. 
	
	\subsubsection{Recognition}
	
	In the recognition step, SPD-ML \cite{18} used the affine-invariant metric and the Stein divergence metric to measure the similarities among SPD matrices. LEML \cite{7} used LEM, under which the SPD matrix Lie group is locally isometric to Lie algebra. LPP and PCA used EM to measure the similarities among vector-form data points. For Lie-LPP, we also use LEM to compute the geodesic distances between two SPD matrices. Unlike the other two algorithms, our algorithm aims to construct a Laplace?Beltrami operator on the SPD matrix Lie group and then learn a more discriminable Lie group, which preserves the geometric and algebraic structure of the original one. 
	
	For YFB DB, we choose $60$ images per subject and run the experiment four times by using each algorithm. In each experiment, we randomly choose $p\left(p=20,30,40,50\right)$ images per subject as the training dataset and the remaining $60-p$ images as the test dataset respectively. For PCA and LPP, we choose the dimension of low dimensional space with $d(d+1)/2$ after dimension reduction, where $d=10$. For Lie-LPP, LEM, and SPD-ML, we choose the low size of the SPD matrix, that is, $d\times d$. Under these choices, the low dimension of PCA and LPP is equal to the low dimension of Lie-LPP. The recognition accuracy results are reported in Table 3. We choose the same classification methods with LEML and SPD-ML in the recognition step. Table 3 shows that the recognition results of Lie-LPP outperform PCA and LPP in any case. Comparisons among Lie-LPP, LEM, and SPD-ML show that Lie-LPP is slightly higher than LEML only when we choose $p = 50$ images per subject. The accuracy recognition rates of Lie-LPP especially outperform the accuracies of LEML and SPD-ML when we choose $p = 20,30,40$. These results show that the effect of dimensionality reduction for SPD matrices by Lie-LPP is better than the effects of LEML and SPD-ML. In addition, LEML and SPD-ML need several parameters when performing their algorithms. Our algorithm only needs to analyze the Laplacian operator of the SPD matrix Lie group and solves the dimensionality reduction problem directly on the SPD matrix Lie group.
	
	\begin{table}[tbp]
		\centering  
		\caption{Classification performance of YFB DB together with the comparison results for Lie-LPP and traditional manifold learning algorithms PCA, LPP, LEML, SPD-ML-Stain, and SPD-ML-Airm. }
		\begin{tabular}{|l|c|c|c|c|}  
			\hline
			YFB DB &YFD-trn20/tst40 &YFD-trn30/tst30 &YFD-trn40/tst20 &YFD-trn50/tst10 \\ \hline
			PCA \cite{19} &$46.4 \pm 1.8$ &$50.2 \pm 2.2$ &$61.8 \pm 1.7$ &$63.7 \pm 1.7$ \\ 
			LPP \cite{5} &$48.2 \pm 1.5$ &$57.2 \pm 2.4$ &$62.8 \pm 1.8$ &$69.4 \pm 1.4$ \\ 
			LEML \cite{7} &$46.5 \pm 1.6$ &$53.6 \pm 2.1$ &$57.3 \pm 1.7$ &$72.9 \pm 1.6$  \\ 
			SPD-Stain \cite{18} &$44.6 \pm 1.9$ &$52.4 \pm 2.1$ & $56.7 \pm 1.3$ &$69.4 \pm 1.7$  \\ 
			SPD-Airm \cite{18} &$45.2 \pm 1.8$ &$53.1 \pm 1.7$ &$58.3 \pm 1.6$ &$68.5 \pm 1.9$ \\
			Lie-LPP &$\textbf{50.8} \pm \textbf{2.4}$ &$\textbf{66.3}\pm \textbf{1.2}$ &$\textbf{70.2} \pm \textbf{1.7}$ &$\textbf{73.9} \pm \textbf{2.1}$ \\\hline
		\end{tabular}	
	\end{table}
	
	
	\section{Conclusions and Future directions}
	
	In summary, the following are the main conclusions of this paper:\\
	\\
	1. 
	\begin{minipage}[t]{0.95\linewidth}
		We construct a Laplace?Beltrami operator on the SPD matrix Lie group and give the corresponding discrete Laplacian matrix.\\
	\end{minipage}\\
	2.
	\begin{minipage}[t]{0.95\linewidth}
		We extend the manifold learning algorithm LPP to Lie-LPP and used it on the SPD Matrix Lie group. We have shown that Lie-LPP can be successfully applied to human action recognition and human face recognition.\\
	\end{minipage}\\
	3.
	\begin{minipage}[t]{0.95\linewidth}
		Our analysis of the geometric and algebraic structure of the SPD matrix Lie group shows that the SPD matrix Lie group is a complete and compact manifold. The sectional curvature on every point is zero; thus, the SPD matrix Lie group is a flat manifold and locally isometric to Lie algebra; \\	 	  
	\end{minipage}\\		 
	4.
	\begin{minipage}[t]{0.95\linewidth}
		However, this is not a simple application of the idea in paper \cite{5}. We developed a new algorithm called Lie-LPP, which is a substantial extension of the LPP algorithm;\\
	\end{minipage}\\
	5.
	\begin{minipage}[t]{0.95\linewidth}
		We analyze the relationships between Lie-LPP and LPP in theory and obtain three theoretical conclusions;\\
	\end{minipage}\\	
	6.
	\begin{minipage}[t]{0.95\linewidth}
		We conducted practical experiments, the results of which show that Lie-LPP outperforms the existing ones significantly.\\
	\end{minipage}\\
	
	In the future, we will further improve this algorithm. We will attempt to add a time dimension to enhance human action recognition, introduce manifold learning algorithms to other types of Lie groups, and introduce new manifold learning algorithms on higher-dimensional tensor space. 
	
	\section*{Acknowledgments}
	
	This work is supported by the National Key Research and Development Program of China under grant no. 2016YFB1000902; the National Natural Science Foundation of China project Nos. 61232015, 61472412, and 61621003; the Beijing Science and Technology Project on Machine Learning-based Stomatology; and the Tsinghua-Tencent-AMSS-Joint Project on WWW Knowledge Structure and its Application.
	
	
	
	
\end{document}